\documentclass{svproc}
\usepackage{graphicx}
\usepackage{hyperref}
\usepackage{url}
\usepackage{float}

\begin{document}
\mainmatter

%
% ---- Title ----
%
\title{Deep Manifold Learning for Reading Comprehension and
Logical Reasoning Tasks with Polytuplet Loss}

\titlerunning{Deep Manifold Learning with Polytuplet Loss}

\author{Jeffrey Lu\inst{1} \and Ivan Rodriguez\inst{2}}

\authorrunning{J. Lu and I. Rodriguez}

\tocauthor{Jeffrey Lu and Ivan Rodriguez}

\institute{York Community High School, Elmhurst IL 60126, USA,\\
\email{jeffrey.p.lu@gmail.com},\\
\and
Brown University, Providence RI 02912, USA,\\
\email{ivan\_felipe\_rodriguez@brown.edu}
}

\maketitle

%
% ---- Main Section ----
%
\begin{abstract}
The current trend in developing machine learning models for reading comprehension and logical reasoning tasks is focused on improving the models’ abilities to understand and utilize logical rules. This work focuses on providing a novel loss function and accompanying model architecture that has more interpretable components than some other models by representing a common strategy employed by humans when given reading comprehension and logical reasoning tasks. Our strategy involves emphasizing relative accuracy over absolute accuracy and can theoretically produce the correct answer with incomplete knowledge. We examine the effectiveness of this strategy to solve reading comprehension and logical reasoning questions. The models were evaluated on the ReClor dataset, a challenging reading comprehension and logical reasoning benchmark. We propose the polytuplet loss function, which forces prioritization of learning the relative correctness of answer choices over learning the true accuracy of each choice. Our results indicate that models employing polytuplet loss outperform existing baseline models, though further research is required to quantify the benefits it may present.
\keywords{contrastive learning, machine learning, natural language processing}
\end{abstract}

\section{Introduction}
Reading comprehension and logical reasoning (RCLR) are critical skills for extracting the most value from written information, which is the primary source of information for state-of-the-art LLMs and many other model classes. RCLR tasks typically involve reading a text sample, reading a question that can be answered based on evidence or inferences from the text, and selecting an answer that can be most reasonably derived from the text and most directly answers the question. The typical approach is prompt-first, placing the greatest focus on the context and question to determine the true accuracy of each answer. However, some human strategies for RCLR tasks suggest that using an alternative approach may result in better performance. This alternative method is answer-first, placing the greatest focus on how the answers compare to each other, nearly disregarding the degree of understanding of the context and question.

An intuitive argument for this alternative approach can be built on its lower threshold for determining the correct answer choice. Consider the following reasoning: Apples are vegetables because all apples are purple, and all purple objects are vegetables. Even without any knowledge of vegetables, one could identify that the reasoning is invalid because the statement “all apples are purple” is false. In this case, only a partial understanding of the information in the prompt is required to correctly determine the accuracy of a statement. More generally, the information sufficient to eliminate incorrect answers is a subset of the information sufficient to determine the absolute accuracy of an answer.

We elected to evaluate our models on ReClor \cite{Yu2020ReClor:}, a challenging RCLR dataset composed of multiple-choice questions with accompanying contexts. In this paper, we compare the performance of models using the alternative approach and traditional baseline models to solve questions presented in ReClor. All models are partially pretrained models, providing the necessary fundamental knowledge to interpret the text data. We propose polytuplet loss, an extended version of triplet loss \cite{SchroffFacenet}, to represent the alternative approach.

\section{Related Work}
Existing approaches to solving reading comprehension with logical reasoning problems tend to focus on the logical reasoning aspect. One such approach is LReasoner \cite{wang-etal-2022-logic}. The fundamental challenge is that language models focus on understanding language rather than reasoning about that understanding. LReasoner used data augmentation to explicitly enforce a logical interpretation and understanding of text over the relatively superficial semantic understanding. Data augmentation has also seen success in other techniques.

As of writing this paper, the most successful documented method is MERIt \cite{jiao-etal-2022-merit}, which introduced a meta-path and counterfactual data augmentation strategy. This approach goes even further to directly create entity-level paths to ensure the model has an understanding of the relationship between important elements of the text. Both LReasoner and MERIt implement contrastive learning \cite{hadsellDimred} for pretraining, which takes advantage of their techniques for data augmentation to force models to learn logical reasoning rather than relying on the information shortcut. In these cases, the primary goal of using contrastive learning was to teach the model to differentiate between semantically similar but logically different answer choices.

\section{Methods}
In this paper, we examine model performance in multiple choice question answering (MCQA) data. With contexts $C$, questions $Q$, answer choices $A_i = \{A_i^1, \ldots, A_i^N\}$, and the correct answer choice $A_i^y$, the goal is to accurately determine $A_i^y$ using $C_i$, $Q_i$, and $A_i$.

We utilized models pretrained on data sourced from books and Wikipedia.
The knowledge is then transferred to classifiers with different loss functions, which are finally fine-tuned for the MCQA task. We employed the Keras-tuner implementation of Hyperband \cite{liHyper} to optimize hyperparameters.

\subsection{Preliminaries}
Our method for contrastive learning applies the methods previously used for facial recognition, verification, and clustering to reading comprehension tasks. Specifically, we draw from the triplet loss approach \cite{SchroffFacenet}. The concept behind triplet loss is to simultaneously minimize the distance between positive and anchor embeddings and maximize the distance between negative and anchor embeddings, but only actively enforcing this until a certain margin $\alpha$ is reached.

With $\alpha$ being the maximum acceptable margin between the positive and negative embeddings and $T$ being the set of all possible triplets of embeddings, each with an anchor embedding $T_i^a$, positive embedding $T_i^p$, and a negative embedding $T_i^n$, triplet loss is defined as follows:

\begin{eqnarray*}
  L_{tri} &=& \sum_{i=1}^{|T|}{ReLU}\left(\left|\left|T_i^a - T_i^p\right|\right|^2_2 - \left|\left|T_i^a - T_i^n\right|\right|^2_2 + \alpha\right)
\end{eqnarray*}
This is minimized when the difference in distances from each negative embedding to the anchor sufficiently distinguishes the negative from the positive.

\subsection{Polytuplet Loss}
We propose a novel loss function that accomplishes the primary goal of contrastive learning \cite{hadsellDimred}, which is to learn transformations from input vectors with high dimensionality to embeddings in a manifold with lower dimensionality such that similar inputs are closer together and dissimilar inputs are further apart.

\label{fig:1}
\begin{figure}[H]
\centering
\includegraphics[width=0.6\textwidth]{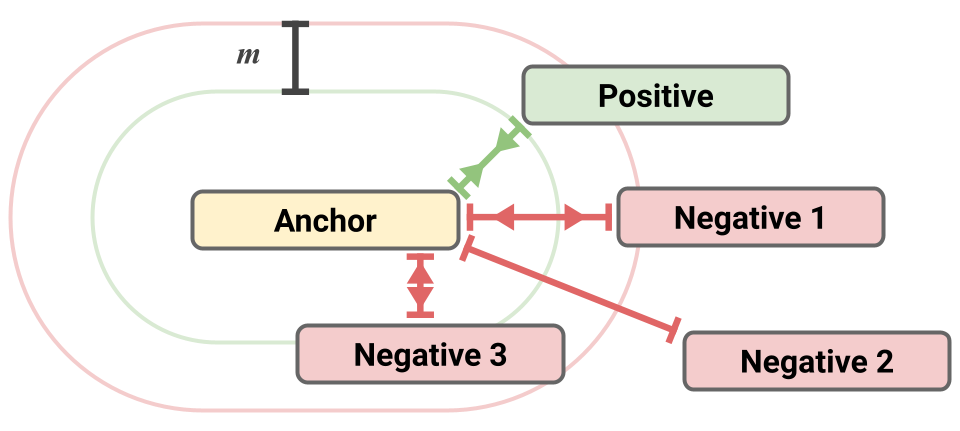}
\caption{The negatives that are not beyond the margin $m$ (1 and 3) being repelled from the Anchor and the Positive being attracted to the Anchor.}
\end{figure}

Projecting embeddings to a manifold restricts the relative scale of distances between embeddings, which is critical to preventing the explosion or collapse of embedding magnitudes. For a transformation $\{C_i, Q_i, A_i^j\} \rightarrow E_i^j$ of a given context, question, and answer to an embedding $E_i^j$ and the transformation $\{C_i, Q_i\} \rightarrow E_i^a$ of a context and question alone to a comparable embedding $E_i^a$, we define \textit{polytuplet loss}, an extension of triplet loss:
\begin{eqnarray*}
  L_{poly} &=& \sum_{i=1}^{|E|}\sum_{j\in\{1, \ldots, N\}, j\neq y} ReLU \left(||E_i^a - E_i^y||^2_2 - ||E_i^a - E_i^j||^2_2 + m\right)
\end{eqnarray*}

Note that this loss, like triplet loss, requires an additional hyperparameter, which we denote $m$ for margin, which is the minimum acceptable margin between the correct answer and incorrect answers relative to the context. Our usage of $m$ is nearly identical to the usage of $\alpha$ in triplet loss.
\nocite{*}

\subsection{Dataset}
The dataset we used was the ReClor \cite{Yu2020ReClor:} training and validation sets, which contain 5138 samples. Each data point has a context, a question, and 4 answer choices. Exactly 1 answer choice is correct; each of the other choices is either logically or factually inaccurate. The ReClor questions are sourced from past GMAT and LSAT exams and high-quality practice exams for these exams.

\label{table:1}
\begin{table}
\begin{center}
\begin{tabular}{r@{\quad\quad}c@{\quad\quad}c@{\quad\quad}cl}
\hline
$y$ & Full Count & Train Count & Test Count &\\
\hline\rule{0pt}{12pt}
0  & 1,299 & 1171 & 129 & \\
1  & 1,288 & 1170 & 129 & \\
2  & 1,283 & 1154 & 125 & \\
3  & 1,268 & 1143 & 117 & \\
\hline\rule{0pt}{12pt}
Total  & 5138 & 4638 & 500 & \\[2pt]
\hline
\end{tabular}
\end{center}
\caption{Distribution of the correct index y in all data}
\end{table}

\subsection{Pretrained Models}

Our pretrained tokenizers and models were from Hugging Face Transformers \cite{wolf-etal-2020-transformers}, which were trained on Book Corpus \cite{ZhuMovies} and English Wikipedia. We selected 4 pretrained models to test based on performance and efficiency: ALBERT, BERT, RoBERTa, and DistilBERT.

\subsection{Baseline Models}
Our baseline models adopted the multiple-choice model architecture from Hugging Face Transformers with an additional softmax activation layer to constrain model outputs to a vector of probabilities. These models were fine-tuned on the ReClor data using the categorical cross-entropy loss. Because each answer choice is evaluated for accuracy independently of others, the baseline models represent the “forward approach” with a higher information threshold.

\subsection{Polytuplet Model}
The polytuplet model takes the same inputs as the baseline model, which we refer to as results, with an additional context. The contexts and results are separately tokenized using an existing pretrained tokenizer and are passed into individually trainable pretrained models. The outputs of the pretrained models are processed using a dense neural network. The embedding layer produces vectors on a unit hypersphere manifold, reducing dimensionality for each context and result. The distances between the context and results on the manifold are used to determine the answer that is the closest to the context, which is predicted as the correct answer. Polytuplet loss is computed using the embedding output of the reduced vector layer. An additional categorical cross-entropy loss is added to the final classification.

\label{fig:2}
\begin{figure}[ht]
\centering
\includegraphics[width=1\textwidth]{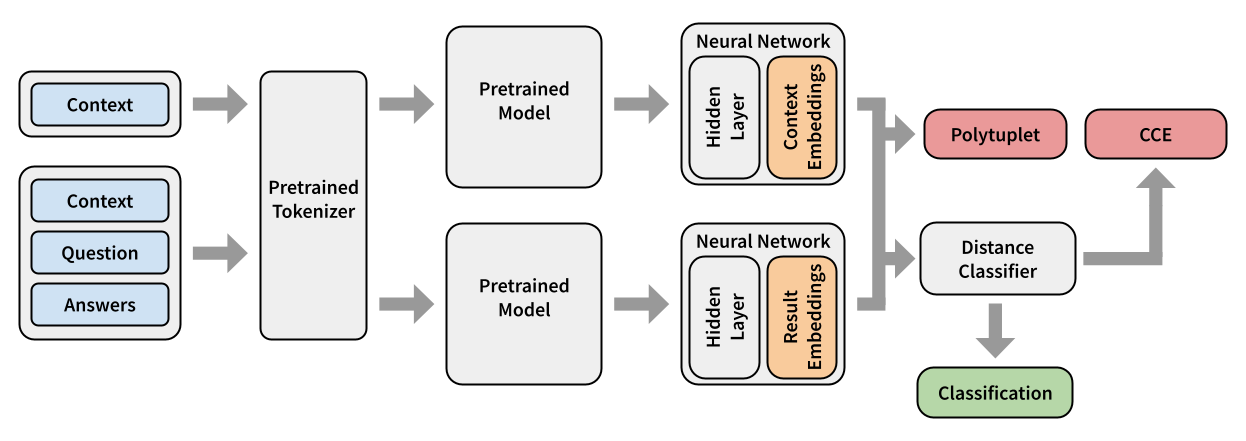}
\caption{Diagram of polytuplet and categorical cross-entropy hybrid model architecture.}
\end{figure}

Essentially, the contexts and results are each processed using pretrained models and converted into embeddings by a dense neural network head. The distances between result embeddings and context embedding are the sole determinants of the model output. Polytuplet loss is computed using the embeddings and categorical cross-entropy loss is computed using the classifications.

\subsection{Index-blind Architecture}

Because of the immense quantity of parameters in language models and the relatively small size of our dataset, overfitting is a significant obstacle to learning the true patterns in the text. To prevent this, we process each result individually and recombine them after projection. This is accomplished using an “index-blind” architecture, which involves flattening the answer choice and batch size dimensions. Specifically, the pretrained model receives $batch\_size \times N$ samples instead of $batch\_size$ samples with $N$ ordered lists of results.

By treating each answer choice as its own independent sample, we aim to eliminate the chance of the model simply memorizing the index of the correct answer of each combination of result embeddings instead of learning meaningful embeddings.

\label{fig:3}
\begin{figure}[ht]
\centering
\includegraphics[width=1\textwidth]{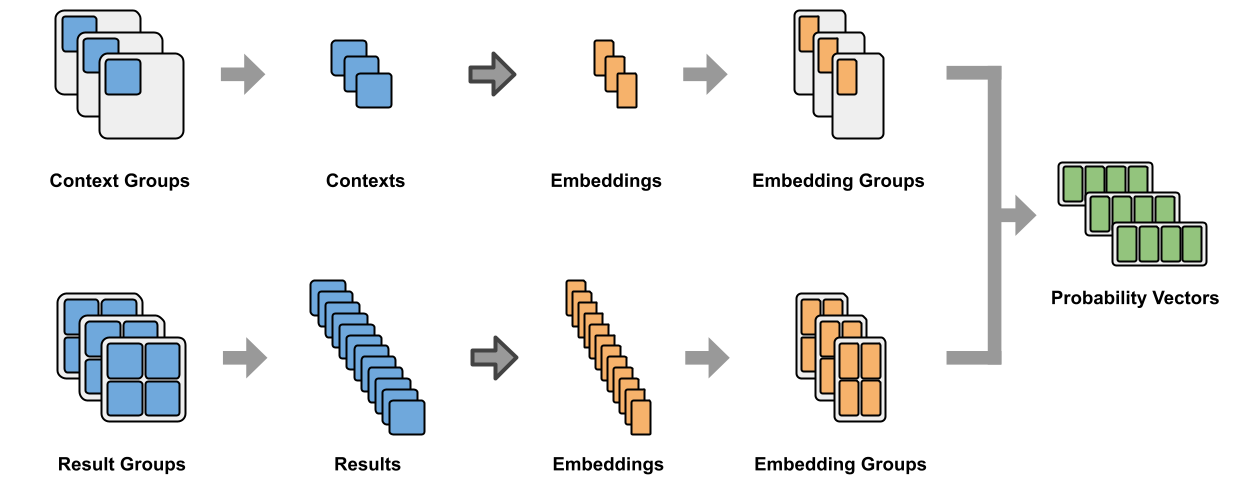}
\caption{The index-blind architecture ensures the point when the model converts tokenized contexts and results to embeddings (highlighted arrows) has no information about the relation between adjacent inputs. Models using polytuplet loss use all paths represented above. Baseline models use only the result pathway (lower) because they do not compare results to contexts.}
\end{figure}

\subsection{Hard and Semi-Hard Negative Mining}
We implemented hard and semi-hard negative mining \cite{SchroffFacenet} to improve the quality of convergence. We classify hard and semi-hard samples nearly identically as they were originally defined in \cite{SchroffFacenet}: hard samples have positive embeddings that are further from the context than at least one negative embedding (See negative 3 in \hyperref[fig:1]{Figure 1}) and semi-hard samples have positive embeddings that are closer to the context than all negative embeddings, but the margin by which they are closer is within the threshold m (See negative 1 in \hyperref[fig:1]{Figure 1}). Specifically, hard negative samples satisfy
\begin{eqnarray*}
  ||E_i^a - E_i^y||_2^2 - ||E_i^a - E_i^j||_2^2 < 0
\end{eqnarray*}
and semi-hard negative samples satisfy
\begin{eqnarray*}
  0 < ||E_i^a - E_i^y||_2^2 - ||E_i^a - E_i^j||_2^2 < m
\end{eqnarray*}
Additionally, we implement the weighting of loss generated from hard and semi-hard negative samples as a tunable hyperparameter, allowing for the optimization of the distribution of hard and semi-hard negative losses in the total loss of the model.

\subsection{Limitations}

Our original experiments were severely limited by the lack of computing resources, including raw power and memory. This restricted our pretrained backbone model types to relatively small and fast pretrained models, which significantly underperform state-of-the-art models. We were later able to access more powerful computing resources to obtain more data on polytuplet performance at greater model sizes. Still, hardware limitations prevented experiments on even larger models that other works used, such as DeBERTa.

Further, our analysis did not consider logic-targeting pretraining or take advantage of the data augmentation and contrastive learning techniques employed by Wang et al. \cite{wang-etal-2022-logic} and Jiao et al. \cite{jiao-etal-2022-merit} because we focus on evaluating the effect of implementing polytuplet loss and its architecture on model performance.

\section{Results and Discussion}

\subsection{Overall Results}
The overall results are shown in \hyperref[table:2]{Table 2}. It can be observed that the polytuplet models outperform the baseline models. This strong performance supports our initial intuition and justification of the alternative approach to RCLR tasks.

\label{table:2}
\begin{table}
\begin{center}
\begin{tabular}{r@{\quad}c@{\quad}c@{\quad}cl}
\hline
 & ALBERT (base) & BERT (base) & DistilBERT (base) &\\
\hline\rule{0pt}{12pt}
Acc. (CCE)  & 53.4 & 44.6 & 44.8 & \\
Acc. (CCE + P)  & \textbf{56.4} & \textbf{49.8} & \textbf{47.7} & \\
\hline\rule{0pt}{12pt}
\% Improvement  & 5.6 & 11.7 & 6.5 & \\[2pt]
\hline
\end{tabular}
\end{center}
\caption{Testing accuracy using CCE (categorical cross-entropy) and CEE + P (categorical cross-entropy and polytuplet) in ALBERT (albert-base-v2), BERT (bert-base-uncased), and DistilBERT (distilbert-base-uncased) based models.}
\end{table}

Between the selected pretrained models, ALBERT performs the best, RoBERTA second, and DistilBERT third. This is expected because ALBERT is known to perform better than the other pretrained models \cite{jiao-etal-2022-merit}.

\subsection{Overfitting}
All models showed signs of overfitting. As noted previously, this is likely due to the models' high complexity and small training dataset size. We implemented dropout \cite{SrivDropout} to reduce the effects of overfitting on model generalization. Tuning the dropout value reveals that polytuplet-based models have optimal performance at lower dropout rates than baseline models, though this trend becomes much less apparent at greater model complexities and has a relatively small effect on the impact of overfitting.

\subsection{Greater Model Complexities}
Because the results on models with lower complexity (\hyperref[table:2]{Table 2}) show strong improvements on baseline models, we tested the effectiveness of polytuplet loss in improving the performance of models with higher complexity.

The results show that on both larger ALBERT and RoBERTa models, the addition of polytuplet loss to training measurably increases performance. This indicates that polytuplet loss may provide meaningful improvements to the 

\label{table:3}
\begin{table}
\begin{center}
\begin{tabular}{r@{\quad}c@{\quad}cl}
\hline
 & ALBERT (xxlarge) & RoBERTa (large) &\\
\hline\rule{0pt}{12pt}
Acc. (CCE)  & 62.6 & 55.6 & \\
Acc. (CCE + P)  & \textbf{67.5} & \textbf{59.1} & \\
\hline\rule{0pt}{12pt}
\% Improvement  & 7.8 & 6.3 & \\[2pt]
\hline
\end{tabular}
\end{center}
\caption{Testing accuracy using CCE (categorical cross-entropy) and CEE + P (categorical cross-entropy and polytuplet) in ALBERT (albert-xxlarge-v2), RoBERTa (roberta-large) based models. Baseline accuracies were obtained from the public leaderboard hosted by Yu et al. \cite{Yu2020ReClor:}}
\end{table}

\section{Conclusion and Future Work}

In this paper, we proposed the polytuplet loss function and an accompanying model architecture for reading comprehension and logical reasoning tasks. Models built on pretrained ALBERT, BERT, RoBERTa, and DistilBERT backbones were evaluated using ReClor, a challenging reading comprehension and logical reasoning dataset. When compared to baseline models, polytuplet models outperformed the baseline models by 5.6-11.7\%. The polytuplet loss is a promising alternative to existing approaches using only categorical cross-entropy for training. Further investigation is needed to compare polytuplet loss to state-of-the-art contrastive learning loss functions for pretraining. Finally, we found that polytuplet-based models sometimes perform better at lower dropout rates, which may indicate that the loss function and architecture are inherently more resistant to overfitting. Again, further work is required to draw significant conclusions. Overall, the answer-focused polytuplet models are a promising, though not definitively superior, new approach to solving reading comprehension and logical reasoning tasks in NLP.

\section{Acknowledgements}

This work was supported by Cloud TPUs from Google's TPU Research Cloud (TRC), which provided access to TPUs and TPU virtual machines. We acknowledge the cloud computing environments made available through Google Colaboratory. This work was also supported and partially funded by Inspirit AI.

This work was completed with the mentorship of Ivan Rodriguez, who provided significant consultations regarding ideation, code design, and analysis. The code and manuscript were written by Jeffrey Lu. Figures were designed by Jeffrey Lu.

%
% ---- Bibliography ----
%
\bibliographystyle{bibtex/splncs03}
\bibliography{refs}

\begin{thebibliography}{1}
\providecommand{\url}[1]{\texttt{#1}}
\providecommand{\urlprefix}{URL }

\bibitem{hadsellDimred}
Hadsell, R., Chopra, S., LeCun, Y.: Dimensionality reduction by learning an invariant mapping. In: 2006 IEEE Computer Society Conference on Computer Vision and Pattern Recognition (CVPR'06). vol.~2, pp. 1735--1742 (2006)

\bibitem{jiao-etal-2022-merit}
Jiao, F., Guo, Y., Song, X., Nie, L.: {MERI}t: {M}eta-{P}ath {G}uided {C}ontrastive {L}earning for {L}ogical {R}easoning. In: Findings of the Association for Computational Linguistics: ACL 2022. pp. 3496--3509. Association for Computational Linguistics, Dublin, Ireland (May 2022), \url{https://aclanthology.org/2022.findings-acl.276}

\bibitem{liHyper}
Li, L., Jamieson, K., DeSalvo, G., Rostamizadeh, A., Talwalkar, A.: Hyperband: A novel bandit-based approach to hyperparameter optimization. J. Mach. Learn. Res.  18(1),  6765–6816 (jan 2017)

\bibitem{SchroffFacenet}
Schroff, F., Kalenichenko, D., Philbin, J.: Facenet: A unified embedding for face recognition and clustering. In: 2015 IEEE Conference on Computer Vision and Pattern Recognition (CVPR). pp. 815--823 (2015)

\bibitem{SrivDropout}
Srivastava, N., Hinton, G., Krizhevsky, A., Sutskever, I., Salakhutdinov, R.: Dropout: A simple way to prevent neural networks from overfitting. J. Mach. Learn. Res.  15(1),  1929–1958 (jan 2014)

\bibitem{wang-etal-2022-logic}
Wang, S., Zhong, W., Tang, D., Wei, Z., Fan, Z., Jiang, D., Zhou, M., Duan, N.: Logic-driven context extension and data augmentation for logical reasoning of text. In: Findings of the Association for Computational Linguistics: ACL 2022. pp. 1619--1629. Association for Computational Linguistics, Dublin, Ireland (May 2022), \url{https://aclanthology.org/2022.findings-acl.127}

\bibitem{wolf-etal-2020-transformers}
Wolf, T., Debut, L., Sanh, V., Chaumond, J., Delangue, C., Moi, A., Cistac, P., Rault, T., Louf, R., Funtowicz, M., Davison, J., Shleifer, S., von Platen, P., Ma, C., Jernite, Y., Plu, J., Xu, C., Le~Scao, T., Gugger, S., Drame, M., Lhoest, Q., Rush, A.: Transformers: State-of-the-art natural language processing. In: Proceedings of the 2020 Conference on Empirical Methods in Natural Language Processing: System Demonstrations. pp. 38--45. Association for Computational Linguistics, Online (Oct 2020), \url{https://aclanthology.org/2020.emnlp-demos.6}

\bibitem{Yu2020ReClor:}
Yu, W., Jiang, Z., Dong, Y., Feng, J.: Reclor: A reading comprehension dataset requiring logical reasoning. In: International Conference on Learning Representations (2020), \url{https://openreview.net/forum?id=HJgJtT4tvB}

\bibitem{ZhuMovies}
Zhu, Y., Kiros, R., Zemel, R., Salakhutdinov, R., Urtasun, R., Torralba, A., Fidler, S.: Aligning books and movies: Towards story-like visual explanations by watching movies and reading books. In: 2015 IEEE International Conference on Computer Vision (ICCV). pp. 19--27. IEEE Computer Society, Los Alamitos, CA, USA (dec 2015), \url{https://doi.ieeecomputersociety.org/10.1109/ICCV.2015.11}

\end{thebibliography}

\end{document}